# A Novel Scheme for Generating Secure Face Templates Using BDA

Shraddha S. Shinde
P.G. Student, Department of Computer Engineering,
MCERC,
Nashik (M.S.), India
e-mail: shraddhashinde@gmail.com

Prof. Anagha P. Khedkar
Associate Professor, Department of Computer Engineering,
MCERC, Nashik (M.S.), India
e-mail: anagha_p2@yahoo.com

*Abstract—* In identity management system, frequently used biometric recognition system needs awareness towards issue of protecting biometric template as far as more reliable solution is apprehensive. In sight of this biometric template protection algorithm should gratify the basic requirements viz. security, discriminability and cancelability. As no single template protection method is capable of satisfying these requirements, a novel scheme for face template generation and protection is proposed. The novel scheme is proposed to provide security and accuracy in new user enrolment and authentication process. This novel scheme takes advantage of both the hybrid approach and the binary discriminant analysis algorithm. This algorithm is designed on the basis of random projection, binary discriminant analysis and fuzzy commitment scheme. Publicly available benchmark face databases (FERET, FRGC, CMU-PIE) and other datasets are used for evaluation. The proposed novel scheme enhances the discriminability and recognition accuracy in terms of matching score of the face images for each stage and provides high security against potential attacks namely brute force and smart attacks. In this paper, we discuss results viz. averages matching score, computation time and security for hybrid approach and novel approach.

## I. INTRODUCTION

Biometric systems are being deployed in various applications including banking, airports, health care and border crossing, thus improving security and discriminability of biometric template. Biometric systems used the human face, a feature as a template. At several airports use system which analyze facial features in face recognition to increase security. In spite of many advantages, biometric systems like any other security applications are vulnerable to a wide range of attacks.

Recently, Vietnamese researchers have cracked facial recognition technology in Lenovo, Asus, and Toshiba laptops and demonstrated vulnerabilities in the systems that let an attacker cheat them with phony photos (face brute force attack) of the legitimate user and gain access to the laptops [40][41]. Hill-climbing attack modifies the raw input biometric data iteratively based on the output matching score from the system for accessing the system [38]. Masquerade attack reconstructs a raw biometric template from the template stored in database, which can be used for accessing the system in future. The feature extractor may be attacked with a Trojan horse program that produces predetermined feature sets. Legitimate feature sets extracted from the biometric input may be replaced with synthetic feature sets. The enrolled templates in the database may be modified or removed, or new templates may be introduced in the database instead of the original templates, which could result in authorization for an intruder, or at least denial of service for the person associated with that corrupted or modified template. Spoof attack as stolen, copied or synthetically replicated biometric trait to the sensor to damage the biometric system security in order to gain unauthorized access. An attack on a biometric system can take place for three main reasons [39]:

- A person may wish to disguise his own identity. For instance, an individual/terrorist attempting to enter a country without legal permission may try to modify his biometric trait or conceal it by placing an artificial biometric trait (e.g. a synthetic fingerprint, mask, or contact lens) over his biometric trait. Recently, in January 2009, the Japanese border control fingerprint system was deceived by a woman who used tape-made artificial fingerprints on her true fingerprints.
- If any individual wants to attain the privileges of other person, then an attack on biometric system can occur. The impostor, in this case, may forge biometric trait of genuine user in order to gain the unauthorized access to systems such as person's bank account or to gain physical access to a confidential region.
- A benefit to sharing biometric trait may be the cause to attack the biometric systems. Someone, for instance, can establish a new identity during enrollment using a synthetically generated biometric trait. Thus, sharing the artificial biometric trait leads to sharing that fraudulent identity with multiple people.

To enhance the recognition performance as well as to provide the better security, new system is to be proposed. The proposed system is designed

- To enhance the discriminability of face template by using Binary discriminant analysis.

- To provide the better security to binary template against smart attacks and brute force attack.

## II. EXISTING TEMPLATE PROTECTION SCTEMES

Template protection scheme can be organized into three main approaches: the biometric cryptosystem approach, the transform-based approach and hybrid approach. The basic idea of these approaches is that instead of storing the original template, the transformed/encrypted template which is intended to be more secure, is stored. In case the transformed/encrypted template is stolen or lost, it is computationally hard to reconstruct the original template and to determine the original raw biometric data simply from the transformed/encrypted template.

The error-correcting coding techniques are utilized to handle intra-class variations in the biometric cryptosystem approach. Two popular techniques viz. fuzzy commitment scheme [7] and fuzzy vault scheme [8] are discussed. High level security is provided to template by applying encryption. However, the error-correcting ability of these schemes may not be strong enough to handle large intra-class variations for face images captured. Also, this approach is not designed to be revocable. Finally, the error-correcting coding techniques require input in certain format (e.g., binary strings or integer vectors with limited range), and it is hard to represent every biometric template in this desired format.

In the transform-based approach, a transformed template is generated using a "one-way" transform and the matching is performed in the transformed domain. The transform-based approach has a good cancelability property, but the drawback of this approach is the trade-off between performance and security of the transformed template.

The hybrid approach retains the advantages of both the transform-based approach and biometric cryptosystem approach, and defeats the limitations of individual approaches. Some of the existing biometric template protection schemes provide security to binary template and rest of the other enhances the discriminability of the template but not yet sufficiently mature for large scale deployment; they do not meet the requirements of diversity, revocability, security and high recognition performance. So in order to take the benefits of both approaches while eliminating their limitations, a hybrid approach for face biometric was developed only for the verification process but not for the new user enrolment process [10].

This limitation imposed to develop a new system to generate secure and discriminant face template for new user enrolment process as well as for the verification system.

## III. PROPOSED SYSTEM

A Novel scheme to generate discriminant and secure binary face template by using binary discriminating analysis is proposed. In this novel scheme we apply binary discriminant analysis algorithm to enhance the discriminability of binary template. Finally, encryption algorithm is applied to generate secure template using fuzzy commitment. Two stages (enrollment and authentication) of proposed system follow some steps which are explained in following section.

### A. Random Projection

Random projection is a popular dimensionality reduction technique and has been successfully applied in many computer vision and pattern recognition applications. Recently, it has also been employed as a cancelable transform for face biometric [9] [10]. The main purpose of the original random projection is to project a set of vectors into a lower dimensional subspace. After projecting set of vectors the projection matrix is stored into database into encrypted format.

### B. Binary Discriminant Analysis

BDA algorithm requires minimized within class variance and maximized between class variance. Perceptron method is used to find the optimal linear discriminant function. In the training phase, genuine label for each class is required by the perceptron method. The perceptron minimizes the distance between binary templates to the corresponding target binary template. This binary template is used as the reference for each class.

### C. Fuzzy Commitment

In this stage, to provide better security encryption is applied on binary template. The encrypted binary template is stored into database or to match with stored template.

### D. Algorithm of Proposed System

A novel scheme consist two stages as enrollment and authentication stage. Algorithm for the novel approach is mentioned below.

*1) Step 1: Enrollment*
(a) Input m x n training face templates $T_{pq}$ (p = 1; 2,..k; q = 1, 2,..r) from k classes, each class has r training samples.
(b) Denote l as the length of the training template. Randomly generate a series of ($l_0$ x $l_r$) matrices $M_s$ (s = 1,2,…k), and orthogonalize the columns of $R_s$ with Gram-Schmidt algorithm, where $l_r$ is the length of the generated cancelable template.
(c) Apply binary discriminant analysis on projection vector to generate binary template.
(d) Apply encryption on binary template to create encrypted binary template.

*2) Step 2: Authentication*
(a) A query template is presented with a projection matrix.
(b) Generate a cancelable template.
(c) Apply binary discriminant analysis and generate a binary template.
(d) Release the stored data. Compute encrypted binary template and perform matching for generated template with stored data.
(e) Check the result as acceptance or rejection.

## IV. ARCHTECTURAL DESIGN OF PROPOSED SYSTEM

In this section, we explore the system architecture and its technique to reduce space and time required for the computation. Following sections explain serialization technique and rationale, architecture and performance measurement of proposed system.

### A. Dynamic Programming and Serialization

The proposed system follows the serialization technique. But for the matching of stored template with newly generated template can be achieved by dynamically. Feature extraction, random projection, binary discriminant analysis, encrypt binary template all these steps will execute one by one.

Binary discriminant analysis algorithm is designed to optimize the discriminability of the binary templates. Optimization cannot be directly applied to binarization scheme. So to achieve optimization, perceptron objective function is used by the researchers. This method use a set of n linear discrminant functions to transform a real-valued face template into n dimensional binary template. This approach follows the global binarization scheme; each linear discriminant function transforms the real valued template into a bit. The n-dimensional binary template is constructed by applying all n functions. For this step divide and conquer approach is applied to achieve the speed for calculations.

Finally, instead of storing the images as it is, it is converted into the matrix format which will help to reduce the memory required to store images (matrix format) into the database as well as the computation time will be reduced.

### B. Data Flow Architecture

The proposed system convert the color face image into matrix format and converted matrix will be used to perform the various functions defined in the system. The resultant encrypted matrix will be stored into the database. These stored templates will be used for matching process independently. The following figure 1 shows the data flow architecture for the proposed system.

### C. Rationale of Proposed System

The proposed system use input data independence (matrix format) and serialization for exploring effective use of synchronized processing. Concurrent processing will be applied to construct the n dimensional binary template in the Binary discriminant analysis module. Figure 2 shows the rationale of the binary discriminant analysis algorithm.

### D. Performance Measurements of Proposed System

Here we discuss the performance measuring parameters required for each module of proposed system. Table no. 1 shows the parameters for each module viz. feature vector, vector length, discriminability and security strength to measure the performance of proposed system.

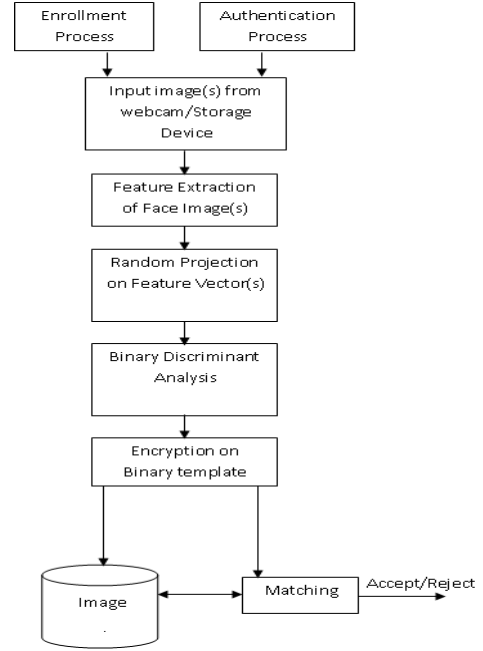

Figure 1. Data flow architecture of proposed system

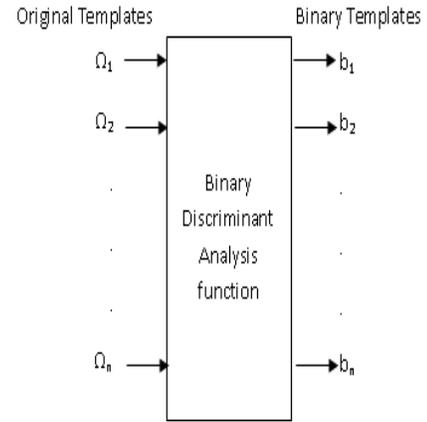

Figure 2. Rationale of proposed system

TABLE I. PERFROMANCE MEASUREMENTS OF PROPOSED SYSTEM

| Input | Modules | Measuring Performance Parameters |
|---|---|---|
| Images | Feature extraction | Face feature vector |
| Face feature vector | Random projection | Vector length(length of the cancellable template), projection matrix for all users (images) |
| Projection matrix | Binary discriminant analysis | Discriminability of the template (in %), length of the binary template |
| Binary template | Encryption | Encrypted matrix, security strength (in bits) |

### E. Datasets used for Proposed System

This section focuses on sample datasets used for testing of proposed system. Three standard benchmark viz. FERET,

CMUPIE, FRGC and other images are shown below (Figure no. 2 to 4). We have selected following face images for testing purpose from available dataset (standard benchmarks are freely available).

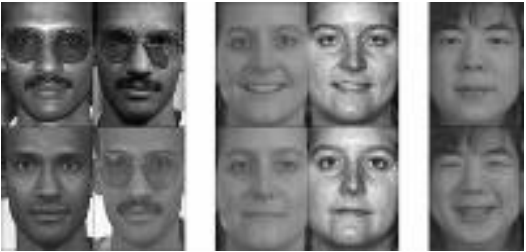

Figure 3. FERET face dataset

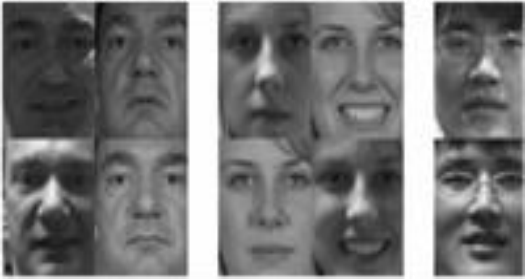

Figure 4. FRGC face dataset

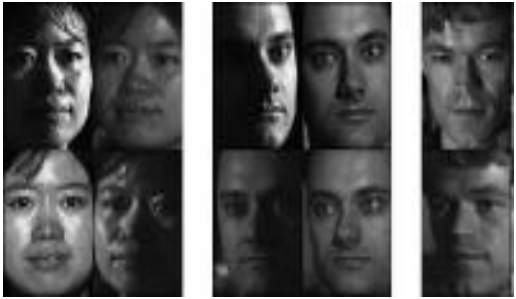

Figure 5. CMUPIE face dataset

## V. RESULTS

This section discusses the results related discriminability and accuracy, computation time and security analysis for the hybrid approach and novel approach.

### A. Discriminability and Accuracy

Three different face templates, namely cancelable template, binary template and secure template are generated by hybrid and novel scheme. In this section we evaluated the discriminability of each template. In Particular, we illustrate that the cancelable template discriminability is enhanced in the binary template and matching score of the templates at each stage of hybrid and novel approach. Table no. 2 shows the average matching score of templates at each stage in hybrid as well as novel approach respectively.

Matching score of feature vector is greater than cancelable template score that means the accuracy and discriminability is degraded in random projection stage. The degraded accuracy and discriminability is enhanced by the binary template generated using DP transform. Matching score of feature vector (from Table 3) is greater than cancelable template score that means the accuracy and discriminability is degraded in random projection stage. The degraded accuracy and discriminability is enhanced by the binary template generated using binary discriminant analysis. Figure 6 shows the graphical representation of these results.

The table no. 4 shows the average matching score for DP transform and binary discriminant analysis. The comparison of both the stages shows matching score for templates generated by binary discriminant analysis is greater than DP transform. Figure 7 shows the graphical representation of this result.

### B. Computation Time

All the experiments are performed on typical personal computer having configuration as core i5 processor and both algorithms (hybrid and novel approach) are implemented using NetBeans IDE. Table no. 5 shows the computation time for one class training process. This time includes random projection, DP Transform and fuzzy commitment for hybrid approach as well for novel approach that includes random projection, BDA process and fuzzy commitment. Here we considered 10 sample images from standard database FERET, FRGC database and other data sets. Analysis of computation time of verification stage is shown in Figure 8.

### C. Security Analysis

This section analyzes the security strength of the hybrid approach, a novel approach at each stage of these algorithms. Two types of potential attacks, namely brute force and "smart" attacks are considered. Brute force attack tries to guess the biometric data without any information, such as matching score, to attack the system. The smart attack viz. affine transformation attack is considered. It can be seen that the security strength of random projection and DP transform are low and medium respectively in smart attack, but full hybrid algorithm is secure against this attack.

In case of novel approach, it can be observed that the security strength of random projection and BDA process are low and medium respectively, but complete novel algorithm is highly secure.

Hybrid approach and a novel approach are highly secure against brute force attack. In this case attacker requires number of trial of all possible combination of all alphanumerical character set. Here we have assumed character set of length 40 (including alphabets, numerical characters and other special characters). The attacker may try to guess the templates of each step.

*1) Random projection*

In hybrid algorithm, T1 is template generated at random projection step having length Kc (Kc is 3772). Therefore, it will cost the attacker $2^{kc-1}$ operations to guess it. So this step is secure against brute force attack.

*2) DP transform*

In this step, T2 is template generated at DP Transform having Kc distinguishing points and there are totally $2^{kc}$ combinations. Therefore, it will cost the attacker $2^{kc}$ combinations to guess it. With known distinguishing points it is hard to implement brute force attack against this step. So this step is secured against brute force attack.

For the affine transformation attack, the real valued template is very hard to be reconstructed from a binary template. Moreover matching score, distinguishing points are not useful in this attack. Therefore, the DP transform is secured against an affine transformation attack.

*3) Binary discriminant analysis*

In this step, T2 is template generated at BDA having Kc length and there are totally $2^{kc-1}$ combinations. Therefore, it will cost the attacker $2^{kc-1}$ combinations to guess it. So this step is secure against brute force attack.

For the affine transformation attack, the real valued template is very had to be reconstructed from a binary template. Moreover matching score is not useful in this attack. Therefore, the BDA process is very secure against an affine transformation attack.

*4) Fuzzy commitment scheme*

B1 is template generated in this step having length Kc (Kc is 11340). Therefore, it will cost the attacker $2^{kc-1}$ operations to guess it. So this step is highly secured against brute force attack. Fuzzy Commitment Scheme performs matching operation between two hashed data.

Because of property of the hash function, the distance between these two hash data will not reveal distance information. Therefore matching score is useless for affine transformation attack. Therefore, the Fuzzy Commitment Scheme is secured against an affine transformation attack.

*5) Full algorithm*

Since the three steps are integrated together to form hybrid algorithm, the attacker cannot get the output from these steps. Thus this algorithm does not reveal any information to attackers. In this algorithm, we have encrypted binary template which cannot be easily accessible to the attackers. The output of each step is combined into one string in specific sequence is arranged. After this, the resultant string is converted into bytes which reduce the string length to store into database.

Finally, this converted template is stored or matched with stored database template. Recovery of this stored template from brute force attack requires $2^{kc-1}$ (Kc is 6810) operations. Therefore, binary template recovery is not possible using affine transformation attack. So this full algorithm has high security strength against both attacks.

As the three steps are integrated together to form novel algorithm, the attacker cannot get the output from these steps. Thus this algorithm does not reveal any information to attackers. In this algorithm, we have encrypted binary template which cannot be easily accessible to the attackers. The output of each step is combined into one string in specific sequence is arranged.

After this, this string is converted into bytes which reduce the string length to store into database. Finally, this converted template is stored or matched with stored database template. Recovery of this stored template from brute force attack requires $2^{kc-1}$ (Kc is 6800) operations.

TABLE II. AVERAGE MATCHING SCORE FOR EACH STEP AND FULL ALGORITHM (NOVEL APPROACH AND HYBRID APPROACH)

| Algorithms | Feature Vector | Cancelable Template | Binary Template | Full Algorithm |
|---|---|---|---|---|
| Hybrid Algorithm | 187.6 | 177.8 | 217 | 200 |
| Novel Algorithm | 187.6 | 175.2 | 220 | 198 |

TABLE III. AVERAGE MATCHING SCORE FOR DP TRANSFORM AND BDA ALGORITHM (NOVEL APPROACH AND HYBRID APPROACH

| Stages of algorithms | Average matching score |
|---|---|
| Binary Template Using DP Transform | 217 |
| Binary Template Using BDA | 220 |

TABLE IV. COMPUTATION TIME FOR VERIFICATION STAGE

| Data Set | Hybrid Approach | A Novel Approach |
|---|---|---|
| FERET | 348 | 335 |
| FRGC | 366 | 370 |
| CMU-PIE | 342 | 340 |
| OTHER | 372.6 | 365 |

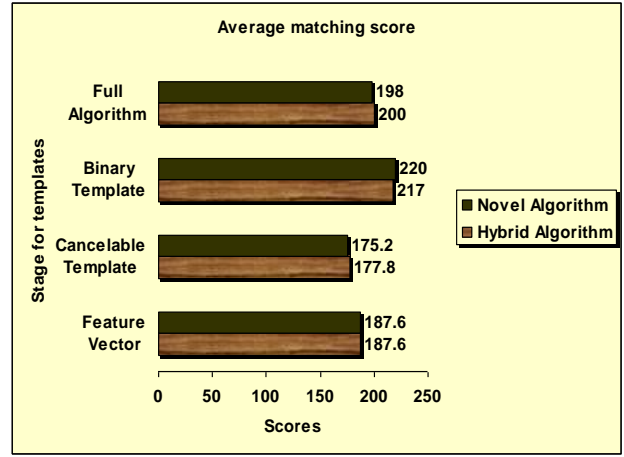

Figure 6. Average matching score for each step and full algorithm (Novel Approach and Hybrid approach)

## VI. CONCLUSION

A novel scheme using BDA was designed, implemented and rigorously tested on the standard benchmark FERET, FRGC and CMU PIE database. Before developing this novel scheme, hybrid approach was also tested on the similar databases. The results of both the methods viz. novel approach and hybrid approach are compared in terms of discriminabilty and security. This paper shows average results for each stage. This clearly indicates the performance progress in novel approach against hybrid approach. The proposed novel technique enhances the discriminability and recognition accuracy in terms of matching score of the face images and provides high security.

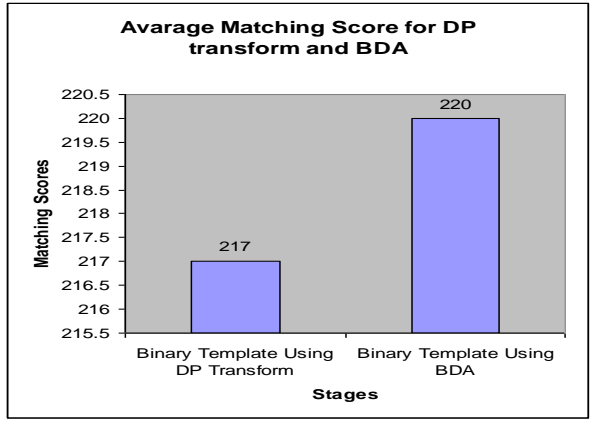

Figure 7.  Average matching score for DP transform and BDA algorithm (Novel Approach and Hybrid approach).

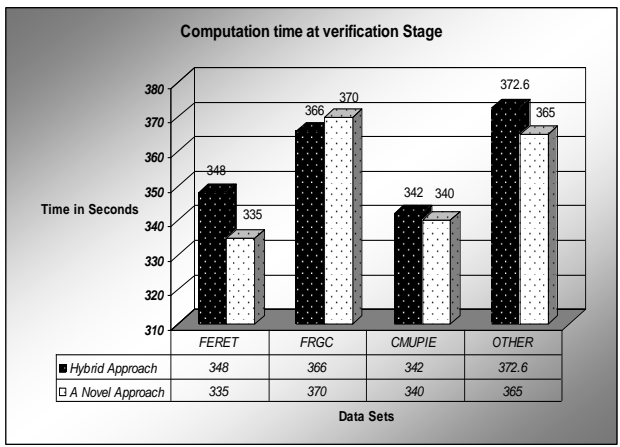

Figure 8.  Computation time for verification stage